\documentclass{article}

\usepackage{arxiv}

\usepackage[utf8]{inputenc} 
\usepackage[T1]{fontenc}    
\usepackage{hyperref}       
\usepackage{url}            
\usepackage{booktabs}       
\usepackage{amsfonts}       
\usepackage{nicefrac}       
\usepackage{microtype}      
\usepackage{lipsum}		
\usepackage{graphicx}
\usepackage{natbib}
\usepackage{doi}

\usepackage{xcolor}         
\usepackage{amsmath} 
\usepackage{amssymb}
\usepackage{colortbl} 
\usepackage{xcolor}
\usepackage{siunitx}
\definecolor{headergray}{gray}{0.9} 
\definecolor{groupgray}{gray}{0.95} 
\definecolor{rowaltgray}{gray}{0.98} 
\usepackage{wrapfig}
\definecolor{blue}{rgb}{0.13,0.13,1}
\usepackage{hyperref}
\usepackage{fontawesome5}

\title{WereWolf-Plus: An Update of Werewolf Game setting Based on DSGBench}

\author{
\normalsize ~\textbf{Xinyuan Xia}
\thanks{All authors contributed to the work equally and should be regarded as co-first authors.}, 
\normalsize \textbf{Yuanyi Song}\footnotemark[1], 
\normalsize ~\textbf{Haomin Ma}\footnotemark[1],
\normalsize \textbf{Jinyu Cai}\footnotemark[1]\\
\normalsize Shanghai Jiao Tong University\\
\normalsize \texttt{char-lotte@sjtu.edu.cn,norsheep919@sjtu.edu.cn}\\
\normalsize \texttt{heath7@sjtu.edu.cn,cai\_jinyu@sjtu.edu.cn} \\
}

\date{}

\renewcommand{\shorttitle}{WereWolf-Plus}

\hypersetup{
pdftitle={A template for the arxiv style},
pdfsubject={q-bio.NC, q-bio.QM},
pdfauthor={David S.~Hippocampus, Elias D.~Striatum},
pdfkeywords={First keyword, Second keyword, More},
}

\begin{document}
\maketitle

\begin{abstract}

With the rapid development of LLM-based agents, increasing attention has been given to their social interaction and strategic reasoning capabilities. However, existing Werewolf-based benchmarking platforms suffer from overly simplified game settings, incomplete evaluation metrics, and poor scalability. To address these limitations, we propose WereWolf-Plus, a multi-model, multi-dimensional, and multi-method benchmarking platform for evaluating multi-agent strategic reasoning in the Werewolf game. The platform offers strong extensibility, supporting customizable configurations for roles such as Seer, Witch, Hunter, Guard, and Sheriff, along with flexible model assignment and reasoning enhancement strategies for different roles. In addition, we introduce a comprehensive set of quantitative evaluation metrics for all special roles, werewolves, and the sheriff, and enrich the assessment dimensions for agent reasoning ability, cooperation capacity, and social influence. WereWolf-Plus provides a more flexible and reliable environment for advancing research on inference and strategic interaction within multi-agent communities. Our code is open sourced at  \href{https://github.com/MinstrelsyXia/WereWolfPlus}{\faGithub~MinstrelsyXia/WereWolfPlus}

\end{abstract}

\section{Introduction}

In recent years, Large Language Models (LLMs) have made remarkable progress in academic intelligence, excelling at tasks such as question answering, code generation, and standardized testing. However, their social intelligence such as the ability to reason, communicate, and act strategically in multi-agent social contexts—remains significantly underexplored. This gap limits our understanding of how LLMs behave in interactive, uncertain, and language-driven environments where cooperation, deception, and persuasion play central roles. Addressing this issue is crucial for advancing LLM-based agents in real-world applications involving negotiation, coordination, and competition.

Werewolf, as a classic social deduction game characterized by partial observability, hidden roles, and language-dominated interaction, offers a natural and challenging environment for studying LLM agents’ behavior in complex multi-agent interactions. Building a reliable platform for evaluating and enhancing LLMs’ strategic capabilities in Werewolf is thus of both academic and practical significance.

Recent years have witnessed preliminary progress in evaluating and enhancing the gaming capabilities of LLM-based agents in Werewolf game environments. The DSGBench framework~\citep{tang2025dsgbench} introduced several key metrics including Identity Recognition Proficiency, Killing Survival Rate, and Voting Selection Score. However, its evaluation dimensions primarily focus on generic player competencies while inadequately addressing the critical aspect of role-specific skill assessment, overlooking the unique role-balancing mechanism and team collaboration dynamics inherent to Werewolf gameplay. Although supporting heterogeneous model configurations across all roles, the framework exhibits significant deficiencies in its role competency evaluation system. The Helmsman study~\citep{du2024helmsman} innovatively leveraged the sheriff mechanism in Werewolf to assess LLMs' social intelligence. Nevertheless, its evaluation metric remains overly simplistic, fails to propose corresponding capability enhancement methods, and only supports heterogeneous model configuration for the sheriff role. The ELLM~\citep{xu2023exploring} research addressed context length limitations of LLMs through retrieval-augmented memory mechanisms and historical information reflection approaches, effectively improving gaming performance. Yet this work similarly lacks a comprehensive evaluation index system and only accommodates homogeneous model configurations.

Through systematic analysis of existing research, this study identifies three critical limitations in current approaches: First, existing evaluation systems demonstrate insufficient comprehensiveness, failing to establish a systematic assessment framework encompassing multidimensional aspects such as role-specific skills, social reasoning, and strategic decision-making. Second, the systems demonstrate poor scalability, primarily evidenced by: inadequate support for fully heterogeneous model configurations; absence of implementation for classic roles (e.g., Witch, Hunter); limited compatibility with standard game setups (8 or 12-player configurations); and underdeveloped mechanisms for flexible player number adjustments. Third, technical implementations suffer from multiple issues including violations of official Werewolf rules, logical and information maintenance errors, poorly modularized code architecture, and limited reusability - all of which substantially compromise their reliability as benchmarking platforms.

To address these limitations, we propose a modular, rule-compliant, and extensible Werewolf simulation framework tailored for LLM agents. Our environment faithfully reconstructs the full game process, supports standard boards and customizable roles, enables flexible and fine-grained LLM-role assignment, and incorporates retrieval-augmented memory to construct compact contexts and enhance reasoning. We also implement a suite of comprehensive evaluation metrics that assess agents across multiple capabilities, including deception, skill use, cooperation, and survival. Table ~\ref{comparison with benchmark} shows a comparison between our work and prior works.   

\begin{table}[tbp]
    \centering
    \caption{Comparison between WereWolf-Plus and prior works}
    \begin{tabular}{c|ccccccc}
                             &  \textbf{Sheriff}            & \textbf{Seer}               & \textbf{Witch}     & \textbf{Hunter}  & \textbf{Guard}    & \textbf{Experience} & \textbf{Multi-oppo}\\
                             \hline
       \textbf{DSGBench}~~\citep{tang2025dsgbench}  &  $\times$     & \checkmark   &  $\times$     &    $\times$   & \checkmark &   $\times$  &  \checkmark  \\
       \textbf{Helmsman}~~\citep{du2024helmsman} & \checkmark  & \checkmark  &  $\times$    &   $\times$ & \checkmark &   $\times$  & \checkmark\\
      \textbf{ELLM}~~\citep{xu2023exploring}            &     $\times$    & \checkmark  &  \checkmark &     $\times$   & \checkmark & \checkmark &  $\times$ \\
       \textbf{Werewolf-Plus}  & \checkmark & \checkmark  & \checkmark  &  \checkmark &  \checkmark  & \checkmark  &      \checkmark\\
    \end{tabular}
    \label{comparison with benchmark}
\end{table}

Our contributions can be summarized as followings:
\begin{itemize}
    \item Built the first open-source Werewolf environment, which strictly follows the rules of the game while supporting flexible characters, including seer, witch, hunter, guard, werewolves, villagers as well as the sheriffs, and customizes the agent model configuration.
    \item Address the input length limitation of LLM by retrieving enhanced memory to achieve contextual compression and reflection, and enhance the LLM agent's strategic reasoning ability through contextual learning based on empirical retrieval.
    \item Provide a unified and scalable evaluation framework with quantifiable metrics customized for each game character and player to analyze the LLM agent's social and inferential intelligence through the Werewolf game system.
\end{itemize}

This work lays the groundwork for rigorous and reproducible research on LLM behavior in complex, language-centric, multi-agent games, and offers a reliable testbed for both capability assessment and strategy enhancement.

\section{Related Work}
\begin{figure}[ht]
    \centering
    \includegraphics[width=1\linewidth]{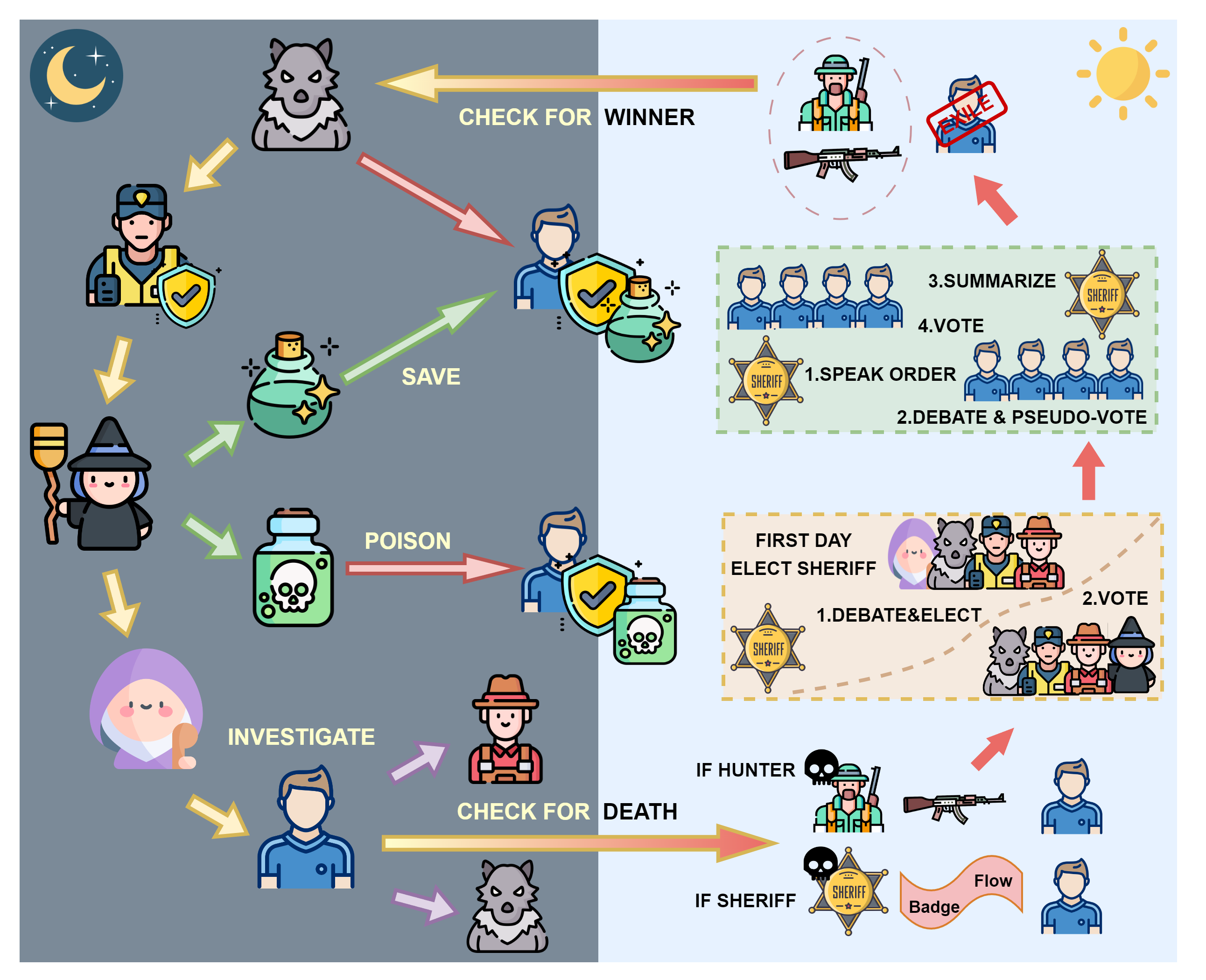}
    \caption{The Complete Game Flow of WereWolf-Plus}
    \label{fig:werewolf-flow}
    \vspace{-10pt}
\end{figure}
\subsection{Game Setup and Flow}
\subsection{DSGBench}

DSGBench (Diverse Strategic Game Benchmark)~~\citep{tang2025dsgbench} is a standardized evaluation platform designed for LLM-based agents in complex strategic decision-making tasks. Nowadays, most of existing benchmarks focus on single objective tasks and use overly broad assessing metrics. Motivated by that, DSGBench integrates six complex strategic games, including StarCraft II, Civilization, Street Fighter III, Werewolf, and Military ChessDiplomacy, spanning real-time and turn-based modes, complete and incomplete information, as well as single-agent competition and multi-agent collaboration. DSGBench introduces fine-grained, multi-dimensional evaluation metrics across five key aspects: strategic planning, real-time decision-making, environmental adaptation, teamwork, and policy stability. Additionally, it incorporates automated decision-tracking mechanism to facilitate behavioral analysis and strategy evolution. This benchmark enables comprehensive assessment and optimization of LLM agents' behavior patterns and changes of strategies, supporting research in reinforcement learning, multi-agent collaboration, and strategic reasoning.

\subsection{Helmsman}

Helmsman is a benchmark designed to evaluate the opinion leadership of large language models (LLMs) within multi-agent decision-making scenarios, using the Werewolf game as a simulation platform~~\citep{du2024helmsman}. In Werewolf, the sheriff is elected by other players and takes on the role of summarizing discussions, proposing decisions, providing information, and influencing outcomes — making it a suitable proxy for opinion leadership. Helmsman introduces a set of metrics to assess the leadership ability of the sheriff, which we adopt in our work. Specifically, a mock vote is conducted after all non-sheriff players have spoken during the daytime phase, followed by the sheriff summarizing the votes and leading a final official vote.

\subsection{ELLM}

Due to the inherent context length limitations of large language models (LLMs), it is often infeasible to retain the full interaction history in complex multi-turn games such as Werewolf. To address this constraint, ELLM~\citep{xu2023exploring} introduces a retrieval-augmented memory mechanism designed to construct compact yet informative context windows for each agent, and leverages reflection over historical dialogues to enhance the reasoning capabilities of LLMs. The proposed framework maintains a history memory pool to retrieve the utterances and incorporates an experience pool which stores past utterances and reflection outputs after each round, and assigns reward scores based on the final game outcome and total number of rounds played.We adopt both the history memory pool and the experience pool mechanisms, but adapt the retrieval strategy in Sec.~\ref{message_pool}. Instead of using fixed predefined questions, we generate a global summary over the accumulated dialogue history, and use this summary as the query to retrieve the most contextually relevant information from the memory pool.

\section{Method}

\label{game}

Participants in the game are referred to as players, and each player may participate in multiple Werewolf games. A single game consists of multiple rounds, with one night and one day phase constituting a complete round. The identity assigned to a player in each game is called their role or character. The flowchart of our werewolf game is shown in the Fig.~\ref{fig:werewolf-flow}.

\subsubsection{Additional Character}
\label{additional character}
In DSGBench’s Werewolf environment, the only special roles (Gods) included are the Seer and the Guard(Doctor), and the rule preventing the Guard from protecting the same player on consecutive nights is not enforced. To better evaluate agents' in-game performance and inference abilities, we additionally incorporated several of the most common roles and corresponding rules from the standard Werewolf game.
\begin{itemize}
    \item Seer: Check the identity of one player each night to determine whether they are a werewolf.
    \item Witch: Possesses one healing potion and one poison. Each night, the witch may choose to save a player from death or eliminate a player; each potion can be used only once.
    \item Hunter: If eliminated during the night or by vote during the day, the hunter may immediately shoot and eliminate one player.
    \item Guard: Protect one player each night from being killed by werewolves, but cannot guard the same player on consecutive nights.
    \item Sheriff: Elected by vote during the first day, holds the right to speak first and has a vote weight of 1.5. After the first round of voting during each daytime phase, the sheriff will summarize the votes and call for a revote.
\end{itemize}

\subsubsection{Game Flow}

At night, the werewolves select a target to eliminate, the guard chooses one player to protect (cannot guard the same player on consecutive nights), and the seer checks the identity of one player to determine if they are a werewolf. The witch is informed of the targeted victim and may choose to use a healing potion to save them or a poison to eliminate another player (each potion can be used only once). The hunter and sheriff have no actions at night.

During the day, the results of the night are announced, and any eliminated or poisoned players leave the game. If the hunter dies, he immediately use his skill to shoot another player. On the first day, players vote to elect a sheriff, who determines the speaking order for daily discussions. After everyone has finished speaking, all players will \textbf{inference the possible identities} of other survival players and vote. The sheriff’s vote counts as 1.5 votes, while all other players have one vote. The sheriff summarize the votes of the first round and call for a \textbf{revote}. In the event of a tie, there will be an opportunity for debate, and the player with the most votes is eliminated.

\begin{wrapfigure}{r}{0.45\textwidth}
  \centering
  \vspace{-2pt}
    \includegraphics[width=0.4\textwidth]{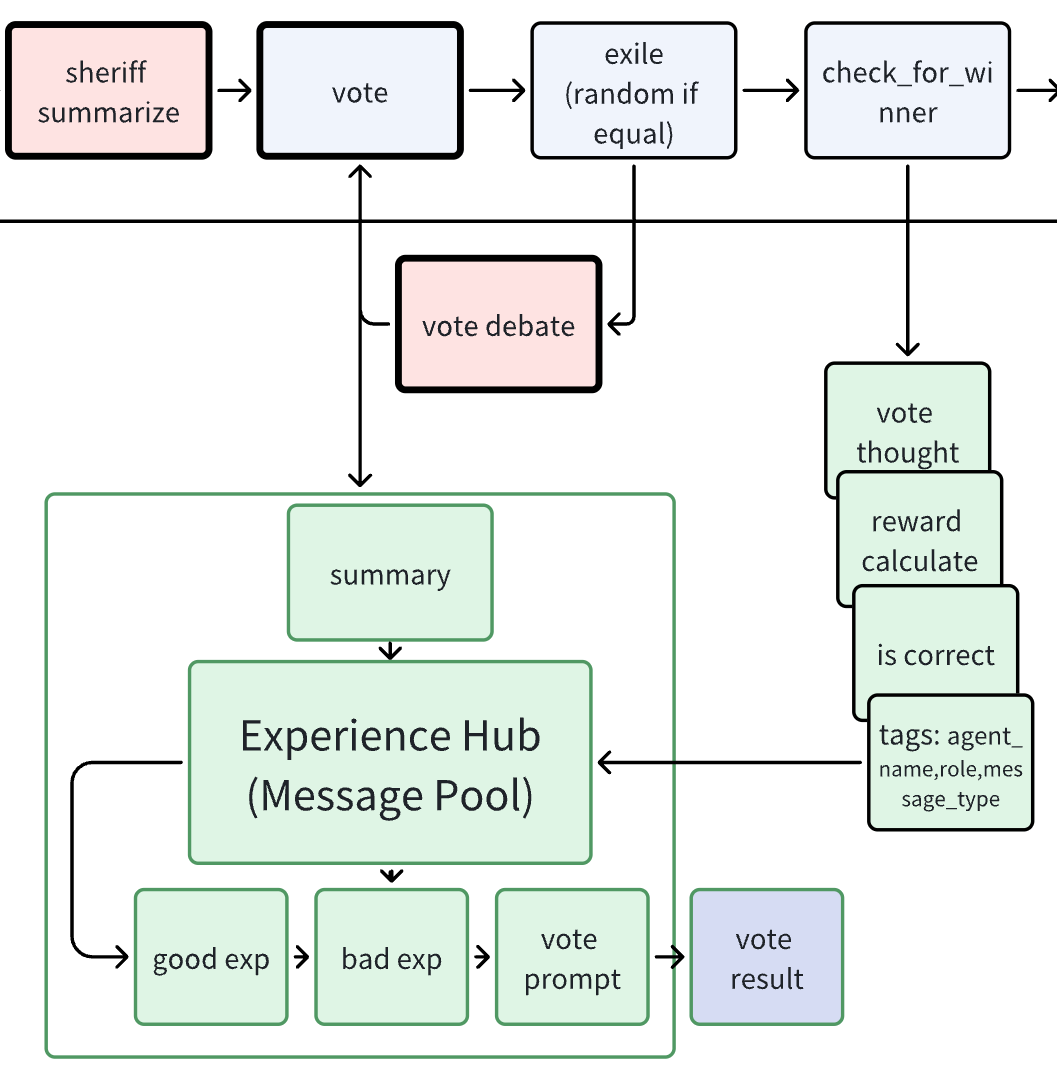}
  \caption{Usage of Vote Experience Pool}
  \label{fig:vote_pool}
  \vspace{-28pt}  
\end{wrapfigure}

An essential term is \textbf{vote change}. Specifically, after the first voting, the sheriff will summarize and analyze the votes as well as game states, recommending an exile player and calling for a new revote. Vote Change is to count the percentage of votes changing to the sheriff's recommendation, which is utilized in Sec.~\ref{evaluation}. 

\subsection{Experience-Retrieval Augmentation}
\label{message_pool}

To enhance the inference capabilities of LLM-based agents, we propose an experience-retrieval augmentation method. Inspired by the self-evolution and message pool implementation of ELLM~~\citep{xu2023exploring}, we design an experience pool that retrieves historical interaction records during game play, enabling context-aware learning and improving agents' reasoning ability through in-game experience reuse. In practice, we only added RAG into voting stage as Fig.~\ref{fig:vote_pool}(not including pseudo-vote), and we put the individual summary about debate before every vote stage into message pool with reward as Eq.~\ref{reward},where $\mathcal{W}$ means win and $\mathcal{L}$ means loss the game. When retrieving the experiences in vote stage, we first calculate the similarity between current summary and history summary in experience pool, filtered by a predefined threshold 0.5, and then choose the top-k experiences with highest reward.
\begin{equation}
\label{reward}
s_i^t = 
\left\{ 
\begin{array}{ll}
1000-T_{max}, & i\in \mathcal{W} \\  
T_{max}, & i\in \mathcal{L}\\  
\end{array}  
\right. 
\end{equation}
For search tags, we support a variety of custom retrieval, allowing for empirical thinking across players, characters, and model ranges.

\subsection{Character-Oriented Evaluation Metrics}
\label{evaluation}

Based on the understanding of the game weredolf, we futher revised some of exsiting player evaluation metrics and designed character-oriented evaluation metics for additional roles.
Here are the the evaluation metrics of key roles, including Gods, Werewolves and the Sheriff.

\begin{itemize}
\label{character score}
    \item \textbf{Seer}: 
    \begin{equation}
    \frac{\# \text{ investigated werewolf}}{\# \text{ total werewolves in game}} 
    \end{equation}
   
    \item \textbf{Witch}: 
    \begin{equation}
    \displaystyle\frac{\# \text{ saved $\in$ good camp \& poisoned $\in$ wolf camp}}{\# \text{ total potion used}}
    \end{equation}
    \item \textbf{Hunter}: 
    \begin{equation}
    \displaystyle\frac{\# \text{ shot $\in$ wolf camp}}{\# \text{ total shot times (constant 1)}}
    \end{equation}
    \item \textbf{Guard}: 
    \begin{equation}
    \displaystyle\frac{\# \text{ $\alpha\cdot($ guarded $\in$ good camp$)+(1-\alpha)\cdot\mathbb{I}($guarded is eliminated$)$}}{\# \text{ total guard attempts}}
    \end{equation}
    \item \textbf{Sheriff}: 
    \begin{equation}
    \displaystyle\frac{\# \text{ change vote $\&$ revote is sheriff's recommendation}}{\# \text{ total vote times}}
    \end{equation}
    \item \textbf{Werewolves}: 
    \begin{equation}
    \displaystyle \alpha\cdot\frac{\# \text{ survival rounds}}{\# \text{ total rounds in one game}}+(1-\alpha)\cdot \mathbb{I}(\text{survived until end})
    \end{equation}
\end{itemize}

\subsection{Player-Oriented Evaluation Metrics}

DSGBench has originally designed some player evaluation metrics for the Werewolf such as Identification Precision (IRP), Key Role Survival Rate (KSR) and Voting Success Score (VSS)~~\citep{tang2025dsgbench}. We have further refined the methods to assess the LLM agent players. 
Here are the evaluation metrics of every player improved on DSGBench by comprehensive performance within multi-games.
\begin{itemize}
    \item Identification Precision (IRP) for Social Reasoning Ability. Quantifies the precision with which a player deduces the roles of other participants, which is same to original version. 
    \begin{equation}
    IRP=\frac{\# \text{ correct identifications}}{\# \text{ total identification attempts}}    
    \end{equation}
    
    \item Key Role Skill Effectiveness (KRE) for Team Collaboration ability. Evaluates the likelihood of key roles surviving until the end of the game and their skill utilization ability defined in Sec.~\ref{character score} for victories of camp.
    \begin{equation}
    KRE=\alpha\cdot\frac{\# \text{ key role survived}}{\# \text{ total key role games}}+ (1-\alpha)\cdot\frac{\# \text{ key role score}}{\# \text{ total key role scores}}
    \end{equation}
    
    \item  Voting Success Score (VSS) for Team Collaboration ability. Assesses the efficacy of a player's voting decisions during pivotal moments in a game of Werewolf.
    \begin{equation}
    VSS=\frac{\# \text{ successful votes}}{\# \text{ total critical votes}}
    \end{equation}
    
\end{itemize}

\section{Experiments}

\subsection{Experimental Setup} 
\textbf{Tasks and Regulations} We played WereWolf games 10 times per configuration. The villager camp wins the game when all werewolves are exiled and the werewolf camp wins when the number of survival werewolves is larger than that of villager camp. We experimented with two models by making them play the good camp and the wolf camp respectively, which becomes a set of experiments.

\textbf{Game Setting} The game is set up with the following angles: the number of players in a round (from 8-12), whether to add the sheriff, whether to add experience pool and who to add.
For game with 8-players, it has 2 werewolves, 4 villagers, and 2 gods respectively seer as well as guard. For game with 12-players, it has 4 werewolves, 4 villagers, and 4 gods respectively seer, witch, hunter as well as guard. 

\textbf{Parameter Configurations} In our implementation, we use following three models: Doubao, DeepSeek-V3, and GPT-4o-mini. We configure the LLM API with a temperature of 1.0, a maximum
token limit of 2048, and a top-p value of 1.0, encouraging the distinction between models and enabling reliable comparisons and rich analysis. For the experience pool experiment, we utilize the "multi-qa-mpnet-base-cos-v1" model to compute the similarity between vote experience queries and database summaries.

\textbf{Main Experiments} We respectively conduct 12-players with no Sheriff and 8-players with Sheriff in order to evaluate the inference and cooperation skills of the agent acting as special characters, and the social skills of the agent acting as sheriff.

\subsection{Main Results}

Table \ref{tab:table1} summarizes the performance of different LLM model agent across different characters and opponents in 12-players game and Table \ref{tab:table3} with no asterisk demonstrates the performance in 8-players game. All metrics range from 0 to 1, and each row records the performance of the LLM model against another LLM model, which is used to measure their relative ability. As shown in the table, Deepseek-V3 performs better than other experimental LLM models in nearly every aspects while GPT-4o-mini shows the worst overall gaming capabilities.

\definecolor{lightblue}{RGB}{230, 240, 255}
\definecolor{headerblue}{RGB}{180,210,255}
\definecolor{rowshade}{RGB}{240,245,255}
\begin{table}[htbp]
  \centering
  \scriptsize \caption{Performance comparison of different models when acting as opponents. We set \textbf{12 players with no Sheriff} for each components and \textbf{10 games} to ensure a fair comparison. The best result is in bold, while the second is underlined. Each two rows represent a set of experiments, with each model acting as good camp and wolf camp.}
  \label{tab:table1}
  \renewcommand{\arraystretch}{0.7}  
  \tiny 
  \resizebox{\textwidth}{!}{
  \begin{tabular}{c|c|c|c|c|c|c|c|c}
    \toprule
    \midrule
    \rowcolor{headerblue}
    \textbf{LLM Model}  & \textbf{Seer}& \textbf{Witch}& \textbf{Hunter}& \textbf{Guard}& \textbf{WereWolf} & \textbf{IRP} & \textbf{KRE} & \textbf{VSS}\\

    \midrule
    {Deepseek-V3} 
    &0.36& 0.41& 1&0.53&0.58&0.83&0.57&0.77\\
    \cmidrule(lr){2-9}
    
    {Doubao} 
    &0.43&0.18&0.25&0.49&0.29&0.46&0.37&0.37 \\
    \midrule
    {Deepseek-V3} 
    &0.39&0.36&0.67&0.53&0.76&0.81&0.77&0.74\\
    \cmidrule(lr){2-9}
    
    {GPT-4o-mini} 
    &0.27&0.18&0.86&0.36&0.32&0.34&0.47&0.18
\\

    \midrule
    {Doubao} 
    &0.41&0.14&0&0.49&0.45&0.89&0.39&0.91\\
    \cmidrule(lr){2-9}
    
    {GPT-4o-mini} 
    &0.43&0.3&0.86&0.28&0.2&0.6&0.43&0.52\\
    \midrule
    \bottomrule
  \end{tabular}
  }
\end{table}

\subsection{Ablation Study on Experience Pool}

\textbf{Experience Pool for Sheriff} In order to verify the impact of experience pools on the sheriff's social influence, we added experience pool in vote stage (not including pseudo-vote) for every players on the basis of 8-players game. In the werewolf game, both good camp and werewolves can become sheriffs, and the Table ~\ref{tab:table3} shows the improvement of both cases. As the table shown, after adding the experience pool, the influence of most models, whether they are good or bad sheriffs, exhibits a certain level of improvement.

\definecolor{lightblue}{RGB}{230, 240, 255}
\definecolor{headerblue}{RGB}{180,210,255}
\definecolor{rowshade}{RGB}{240,245,255}
\begin{table}[ht]
  \centering
  \scriptsize \caption{Performance comparison of different models when acting as opponents. We set \textbf{8 players with Sheriff} for each components and \textbf{10 games} to ensure a fair comparison. Top right corner with `*` means Sheriff influence after adding experience pool. Each two rows represent a set of experiments, with each model acting as good camp and wolf camp.}
  \label{tab:table3}
  \renewcommand{\arraystretch}{0.7}
  \tiny
  \resizebox{\textwidth}{!}{
  \begin{tabular}{c|c|c|c|c|c|c}  
    \toprule
    \midrule
    \rowcolor{headerblue}
    \textbf{LLM Model}  & \textbf{Good Sheriff}&  \textbf{Bad Sheriff} & \textbf{VSS} & \textbf{Good Sheriff*} & \textbf{Bad Sheriff*}  & \textbf{VSS*}\\

    \midrule
    {Deepseek-V3} 
    &0.007&0.053&0.91&0&0.04&0.92\\
    \cmidrule(lr){2-7}
    
    {Doubao} 
    &0 &  0&0.43& 0.093& 0.2& 0.52\\
    \midrule
    {Deepseek-V3} 
    &0.025 &  0&0.65 & 0 & 0&0.84\\
    \cmidrule(lr){2-7}
    
    {GPT-4o-mini} 
    &0.052& 0.032&0.13& 0.034& 0&0\\

    \midrule
    {Doubao} 
    &0.045 &0& 0.54& 0.036& 0&0.58\\
    \cmidrule(lr){2-7}
    
    {GPT-4o-mini} 
    &0.058 & 0& 0.64& 0.074& 0.062&1\\
    \midrule
    \bottomrule
  \end{tabular}
  }
\end{table}

\textbf{Effectiveness of Experiences Pool} To validate the enhancement of reasoning ability using experience pool, we conduct ablation study in Doubao as good camp vs Deepseek-V3 and GPT-4o-mini in 12-players respectively. We only conducted on Doubao because it basic ability is medium and suitable to compare its enhancement. Additionally, We only acted Doubao as good camp in this ablation experiment. For wolf camp, they has already known the roles information so there is no sufficient effect for werewolves to refer history experiences in vote stage. Table ~\ref{tab:table2} demonstrates the enhancements of comprehensive performance after being equipped with vote experience pool.

\definecolor{lightblue}{RGB}{230, 240, 255}
\definecolor{headerblue}{RGB}{180,210,255}
\definecolor{rowshade}{RGB}{240,245,255}
\begin{table}[t]
  \centering
  \scriptsize \caption{Ablation about experience pool on Doubao's performance comparison of different wolf camp opponent models, which shows the inference ability enhancement in \textbf{12 players with no Sheriff} and \textbf{10 games} setup. The best result is in bold.}
  \label{tab:table2}
  \renewcommand{\arraystretch}{0.7}
  \tiny
  \resizebox{\textwidth}{!}{
  \begin{tabular}{cc|c|c|c|c|c|c|c}
    \toprule
    \midrule
    \rowcolor{headerblue}
    \textbf{Setup of Doubao} & \textbf{Opponent Model} & \textbf{Seer}& \textbf{Witch} & \textbf{Hunter}& \textbf{Guard} & \textbf{IRP} & \textbf{KRE} & \textbf{VSS}\\

    \midrule
    {W.O. Experience Pool} 
    & {Deepseek-V3}& 0.43  & 0.18 & 0.25 & 0.49 &  0.46 &  0.37 & 0.37\\
    \cmidrule(lr){2-9}\
    & {GPT-4o-mini} & 0.41  & 0.14 & 0 & 0.49 & \textbf{0.89} &  0.39 & \textbf{0.91}\\
    \midrule
    
    {Vote Experience Pool} 
    & {Deepseek-V3} &0.36&0.18&0.67&  0.40&\textbf{ 0.48} & \textbf{0.44}& \textbf{0.42}\\
    \cmidrule(lr){2-9}\
    & {GPT-4o-mini} & 0.34  & 0.14 & 0.25 & 0.39 & 0.66 & \textbf{ 0.55} & 0.65\\
    \midrule

    \bottomrule
  \end{tabular}
  }
\end{table}

\section{Analysis}

\subsection{Social Capability Analysis on New Quantifiable Metrics}

\textbf{Social Reasoning}
The Identification Precision (IRP) measures the accuracy of a model in inferring the camp of other players, reflecting its social reasoning capability. A higher IRP indicates stronger social inference ability. As shown in Table~\ref{tab:table1}, Deepseek-V3 consistently demonstrates the highest social reasoning ability across different matches, followed by Doubao, with GPT-4o-mini performing the weakest. Notably, in games against Deepseek-V3, the IRP and other performance metrics of the opposing models tend to decline. This suggests that Deepseek-V3 is particularly adept at using language and other cues to deceive opponents and achieve its objectives, exhibiting superior goal-oriented reasoning and social skills.

\textbf{Social Cooperation}
\label{cooperation}
The Key Role Skill Effectiveness (KRE) represents the overall performance score of god roles, while the Voting Success Score (VSS) measures the proportion of successful collective votes against werewolves. By examining KRE and individual role scores, we observe that Deepseek-V3 consistently achieves the highest performance as god roles, along with a high VSS, and also demonstrates the strongest performance when playing as a werewolf. In contrast, both Doubao and GPT-4o-mini show relatively mediocre performance in god roles. However, Doubao achieves notably high VSS in its matches against GPT-4o-mini, suggesting that although its god roles did not effectively utilize their skills to support team objectives, collective reasoning from non-special roles compensated for this gap. Meanwhile, GPT-4o-mini consistently underperforms in VSS across different matches, indicating its limited ability to leverage the advantages of special roles for team victories and reflecting comparatively weaker cooperation capability.

\textbf{Social Influence} The three columns without asterisks in the Table ~\ref{tab:table3} present the Sheriff influence scores in 8-player games, which are used to assess each model’s social influence capability. It can be observed that when aligned with the good camp, GPT-4o-mini demonstrates the strongest influence as Sheriff. In contrast, when assigned as a bad camp Sheriff, Deepseek-V3 achieves the highest influence score, even approaching that of good camp Sheriffs. Combined with its top performance as a werewolf, it suggests that Deepseek-V3 may be particularly skilled at using rhetorical manipulation and ambiguity to mislead the good camp.

\subsection{Inference Enhancement Analysis upon Experience Pool}

The results in Table ~\ref{tab:table3} and Table ~\ref{tab:table2} indicate that the introduction of the experience pool leads to a noticeable improvement for the good camp VSS, confirming the effectiveness of our experience-based enhancement strategy. However, its impact on special roles remains limited. As shown in Table ~\ref{tab:table3}, in 8-player games, the Sheriff’s influence generally increases after incorporating the experience pool. For example, when Doubao serves as the Sheriff against Deepseek-V3, its influence rises to 0.093. Similar improvements are observed in other matches, or in some cases, the overall enhancement in agents’ reasoning ability leads to uniformly accurate voting outcomes, thereby diminishing observable differences in Sheriff influence. Additionally, the VSS improves consistently after integrating the experience pool, suggesting that enhancing contextual reasoning during the voting phase effectively boosts decision-making accuracy. As shown in Table ~\ref{tab:table2}, in Doubao’s match against GPT-4o-mini, both IRP and VSS decrease after adding the experience pool, while the KRE increases. This aligns with our analysis in Sec.~\ref{cooperation}, indicating that Doubao’s special roles did not effectively contribute to overall team success, which may explain the observed performance drop.

\section{Conclusion and Future Work}

In this work, we build upon existing research on multi-agent game evaluation platforms and extend the Werewolf evaluation framework in DSGBench. Our enhanced platform supports standard 12-player games with configurable roles including Seer, Witch, Hunter, and Guard, as well as flexible integration of the Sheriff mechanism. We introduce a comprehensive set of quantitative metrics for evaluating all special roles, werewolves, and the sheriff, while further enriching evaluation dimensions for agent reasoning ability, cooperation, and social influence. The platform also enables customizable role configurations and independent adjustment of reasoning enhancement strategies for different agents, offering a more flexible and reliable benchmarking environment for reasoning and strategic interaction studies in multi-agent communities. Through extensive experiments, we demonstrate the scalability of our platform, the effectiveness of the proposed evaluation metrics, and the performance gains enabled by reasoning enhancement mechanisms.

To further improve agent performance and evaluation comprehensiveness, we plan to extend our work in several directions. (1) Incorporating more baseline models and increasing the number of evaluation rounds to enhance the statistical reliability of experimental results. (2) Assigning different models to special roles, villagers, and werewolves to better capture heterogeneous agent behaviors in strategic interactions. (3) Currently, the experience pool mechanism is applied only during the voting phase; in future work, we will explore extending this mechanism to other decision-making stages such as skill usage and debate, while refining corresponding evaluation metrics to assess its effectiveness. (4) Collecting or constructing additional high-quality game records to enrich the experience pool, and exploring diverse strategy injection methods to further enhance agents’ reasoning ability and adaptability in complex environments.


\bibliographystyle{unsrtnat}
\bibliography{references}  

\newpage
\appendix

\section{Task Division}
Xinyuan Xia: Add message pool, conduct experiments regarding adding message pool, clean the pipeline together with Jinyu Cai, reorganize the open source code.

Yuanyi Song: Add new characters including code, design new assessment metrics, write paper except for introduction, and draw figures.

Haomin Ma: Conduct literature review on prior work, analyze existing code implementations, and write the introduction section of the paper.

Jinyu Cai: Add Sheriff process, code maintenance, conduct experiments of adding roles, fill out the forms.

\section{Ethics Statement}
All experiments in this study are conducted within a simulated virtual environment involving AI agents in Werewolf games, with no involvement of real users or personal data. The research is carried out in a closed simulation setting, posing no impact on actual players or third parties. This work is intended solely to advance research in multi-agent reasoning and strategic decision-making based on large language models, and does not raise any potential ethical concerns or negative social implications.

\section{Cost}

Since we don't have the precise price of our used models and our token usage history through our api key platform, we searched the price on the models' official website and estimated the total cost based on the log we saved. For each model, we calculated the token usage and associated costs as follows:

\begin{enumerate}
    \item GPT-4o-mini: Input price of \$1.10 per million tokens (approximately ¥7.81 per million tokens) and output price of \$4.40 per million tokens (approximately ¥31.24 per million tokens).
    \item Deepseek-v1: Input price of ¥4.0 per million tokens (non-cached) and output price of 16.0 yuan per million tokens. For cached inputs, the price is reduced to 1.0 yuan per million tokens, but we used the non-cached rate for our calculations to provide a conservative estimate.
    \item Doubao: Input price of 5.0 yuan per million tokens and output price of 9.0 yuan per million tokens.
\end{enumerate}

Based on the data above, we estimate that our total cost for conducting our experiments is 179.74 yuan.

\end{document}